\documentclass{article}
\usepackage{spconf,amsmath,graphicx,hyperref}

\usepackage{amssymb}
\usepackage{mathtools}
\usepackage{booktabs}

\usepackage{textcomp}
\usepackage{xcolor}
\usepackage{color}
\usepackage[table]{xcolor}

\usepackage{subcaption}    

\usepackage{amsmath,amsfonts,bm}

\def\eqref#1{equation~\ref{#1}}

\def\1{\bm{1}}

\def\rva{{\mathbf{a}}}

\def\rvx{{\mathbf{x}}}

\DeclareMathAlphabet{\mathsfit}{\encodingdefault}{\sfdefault}{m}{sl}
\SetMathAlphabet{\mathsfit}{bold}{\encodingdefault}{\sfdefault}{bx}{n}

\usepackage{epsfig}
\usepackage{algorithm}
\usepackage{multirow}
\usepackage{graphicx} 

\title{Schr{\"o}Mind: Mitigating Hallucinations in Multimodal Large Language Models via Solving the Schr{\"o}dinger Bridge Problem}

\name{Ziqiang Shi$^{\star}$, Rujie Liu$^{\star}$, 
Shanshan Yu$^{\dagger}$, Satoshi Munakata$^{\dagger}$, Koichi Shirahata$^{\dagger}$}
  
\address{$^{\star}$ Fujitsu Research \& Development Center Co.,LTD., Beijing, China\\
    $^{\dagger}$ Fujitsu Limited, Tokyo, Japan}
\begin{document}
%
\maketitle
\begin{abstract}
Recent advancements in Multimodal Large Language Models (MLLMs) have achieved significant success across various
domains. However, their use in high-stakes fields like health
care remains limited due to persistent hallucinations, where
generated text contradicts or ignores visual input. We contend
that MLLMs can comprehend images but struggle to produce
accurate token sequences. Minor perturbations can shift at
tention from truthful to untruthful states, and the autoregres
sive nature of text generation often prevents error correction.
To address this, we propose SchroMind—a novel framework
reducing hallucinations via solving the Schr{\"o}dinger bridge
problem. It establishes a token-level mapping between hallu
cinatory and truthful activations with minimal transport cost
through lightweight training, while preserving the model's
original capabilities. Extensive experiments on the POPE and
MME benchmarks demonstrate the superiority of Schr{\"o}Mind,
which achieves state-of-the-art performance while introduc
ing only minimal computational overhead.
\end{abstract}
\begin{keywords}
Multimodal large language models, hallucination mitigation, Schr{\"o}dinger bridge problem, 
 entropy-regularized optimal transport, attention activation steering
\end{keywords}
\section{Introduction}
\label{sec:intro}

Multimodal Large Language Models (MLLMs) excel at processing visual and 
textual information, advancing tasks such as visual question 
answering (VQA), image captioning, and visual reasoning \cite{zhu2023minigpt,liu2023visual,wang2024qwen2}. However, they frequently suffer from object hallucination, where 
textual descriptions contradict the image content \cite{leng2024mitigating,huang2024opera,yin2025clearsight,tu2025attention}. This stems from an over-reliance 
on unimodal priors (e.g., linguistic biases) 
during inference, posing significant risks in safety-critical domains like medical diagnosis and autonomous driving.

Existing mitigation strategies often incorporate external knowledge or fine-tune on additional annotated data, but require 
substantial computational resources \cite{liu2024mia,lu2025dama,chen2025perturbollava}. More recent decoding-time interventions, such as contrastive 
decoding \cite{leng2024mitigating,favero2024multi,woo2024dont}, 
suppress hallucinations without retraining, yet may compromise fluency and increase latency. Lightweight inference-time methods that adjust 
activations without updating parameters---known as \textit{activation engineering} \cite{li2023inference,chen2024ict}---are thus gaining attention. Such methods often 
use contrastive samples to derive steering vectors that alter model behavior. For instance, ITI \cite{li2023inference} successfully reduces hallucination in large 
language models (LLMs) by injecting steering activation vectors into attention layers. Subsequent work extended these ideas to MLLMs, leading to methods 
like ICT \cite{chen2024ict}, which operate at global and local visual levels and enhance cross-modal alignment.

\begin{figure}[th]
\begin{center}
   \includegraphics[width=1.0\linewidth]{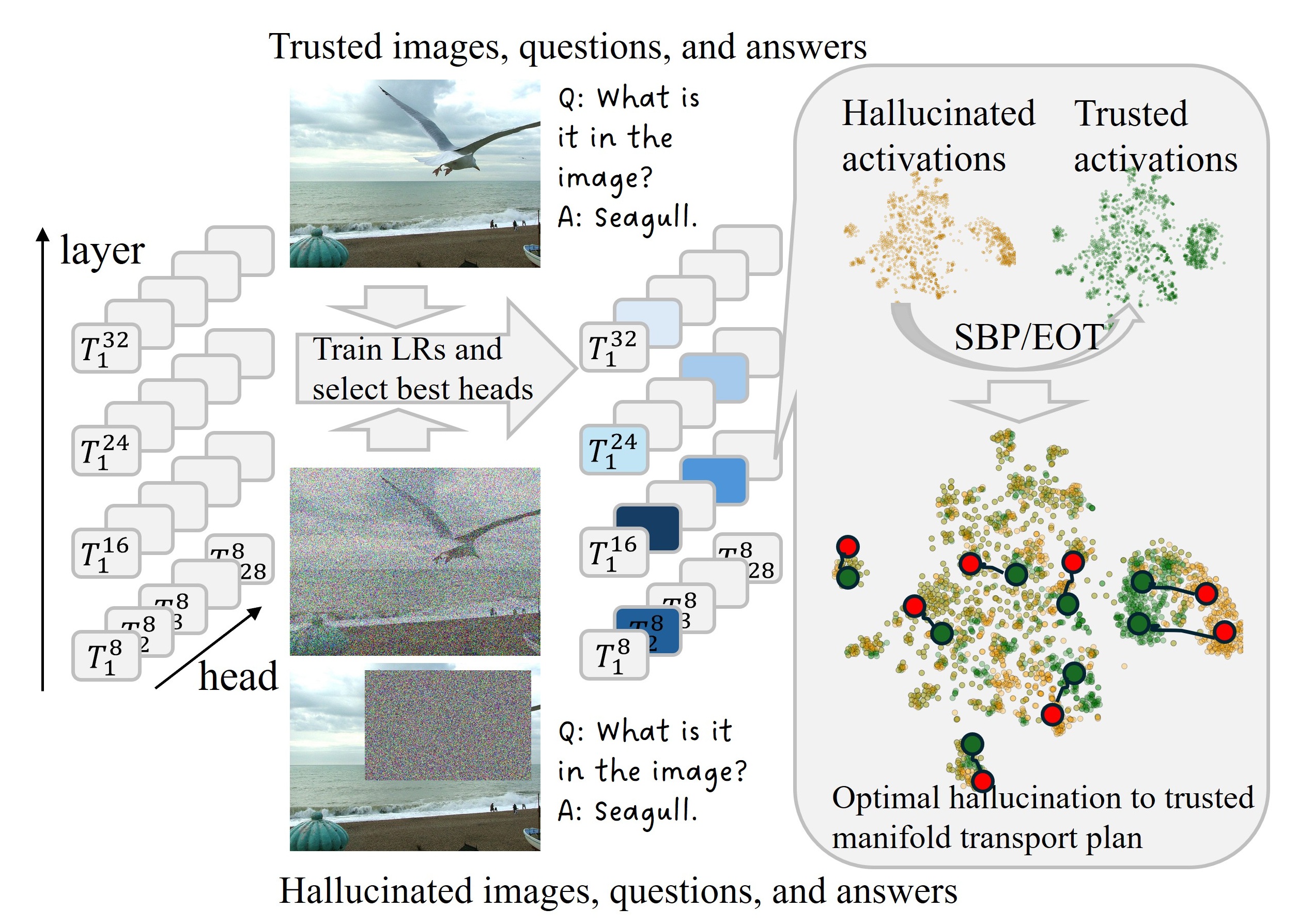}
\end{center}
   \caption{
Schematic of Schr{\"o}Mind. MLLMs process correct and
hallucinated responses to extract attention activations. Classifiers then identify critical attention heads and analyze token
wise influences. SBP/EOT learns a distribution mapping to
mitigate hallucinations.}
\label{fig:framework}
\end{figure}

But most current intervention techniques apply a uniform steering direction across all tokens in an attention head, 
implicitly assuming that shifting the hallucinatory activation manifold yields the truthful manifold \cite{li2023inference,chen2024ict}. We 
argue that this view is oversimplified---each token's activation lies at a unique position, requiring personalized intervention. 
To this end, we propose \textsc{Schr\"{o}Mind}, which models the hallucination and truthful manifolds jointly and derives token-specific 
intervention directions via solving the Schr\"{o}dinger bridge problem (SBP) \cite{leonard2013survey}. This optimal transport-based approach ensures minimal-cost 
mapping between manifolds, enabling precise correction with little training and plug-and-play deployment.

Schr\"{o}Mind's contributions are threefold: 1) We propose a personalized intervention framework that formulates activation correction as an 
optimal transport problem between hallucination and truthful manifolds.  
2) Schr{\"o}Mind is the first to use a Schr{\"o}dinger bridge-based method for token-level 
intervention, requiring only minutes of training and allowing immediate application.  
3) Extensive experiments show that Schr{\"o}Mind outperforms 
previous state-of-the-art (SOTA) methods across multiple benchmarks.

\section{Schr{\"o}Mind}
\label{sec:methodology}

\subsection{Preliminaries and Notation}
\label{sec:notation}

While MLLMs' vision encoders capture sufficient image details, strong language priors in the LLM module can 
induce object hallucinations~\cite{huang2024opera,yin2025clearsight}. To address this, we adjust intermediate 
Transformer states 
toward higher 
reliability. The $K$-layer Transformer is formalized as:
\begin{equation}
\label{eq:transformer}
\rvx^{(k)} = \rvx^{(k-1)} + \sum_{m=1}^{M} 
\mathcal{T}_m^{k-1}\big( \rvx^{(k-1)} \big) \mathbf{\Theta}_m^{k-1},
\end{equation}
where $M$ denotes attention heads per layer, $\mathcal{T}_m^{k-1}(\cdot)$ computes attention-weighted 
features for head $m$ at layer $k-1$, and $\mathbf{\Theta}_m^{k-1} \in \mathbb{R}^{D \times DM}$ projects 
outputs to the $D$-dimensional representation space. The final layer transforms $\rvx^{(K)}$ into vocabulary 
logits, yielding the token prediction distribution:
$$
P(y_t \mid \mathbf{y}_{1:t-1}) = \sigma\Big( \mathcal{L}\big( \rvx_t^{(K)} \big) \Big),
$$
with $\mathbf{y}_{1:t-1}$ as the prior token sequence.

As shown in Fig.~\ref{fig:framework}, we define the hallucinated attention 
distribution $\mathbb{P}_{\text{hallu}} \in \mathcal{P}(\mathbb{R}^D)$ as 
activations induced by hallucinated content due to degraded visual inputs, language 
priors or ignoring vision tokens, whereas the factual 
distribution $\mathbb{P}_{\text{fact}} \in \mathcal{P}(\mathbb{R}^D)$ corresponds 
to authentic visual inputs. Schr\"{o}Mind mitigates hallucinations by 
correcting $\mathbb{P}_{\text{hallu}}$ toward $\mathbb{P}_{\text{fact}}$ via 
solving SBP, 
preserving semantic coherence while aligning distributions through Schr\"{o}dinger 
bridge theory.

\subsection{Defining Hallucination Mitigation in Schr{\"o}Mind}

We construct a mapping transforming hallucinated 
activations $\mathbb{P}_{\text{hallu}}$ into trustworthy 
activations $\mathbb{P}_{\text{fact}}$ using SBP~\cite{leonard2013survey} and 
its static extension, 
the entropy-regularized 
optimal transport (EOT)~\cite{korotin2024light}. 
The SBP/EOT formulation minimizes transport cost with entropy regularization:
\begin{align}
\label{eq:eot}  
\min_{\pi \in \Pi(\mathbb{P}_{\text{hallu}}, \mathbb{P}_{\text{fact}})} 
\iint \frac{1}{2} \| \rva_{\text{hallu}} - \rva_{\text{fact}} \|^2  d\pi
+ \epsilon  \text{KL}(\pi \| \mathbb{P}_{\text{hallu}} \times \mathbb{P}_{\text{fact}}),
\end{align}   
where $\Pi$ denotes valid joint distribution between halluciantion and reliable attention 
activation, the quadratic term measures activation dissimilarity, 
and $\epsilon > 0$ regularizes for stability. The optimal 
plan $\pi^*$ defines a probabilistic correction from hallucinated to factual 
activations.

\begin{figure*}[htbp]
    \centering
    \begin{subfigure}[b]{0.24\linewidth}
        \centering
        \includegraphics[width=\linewidth]{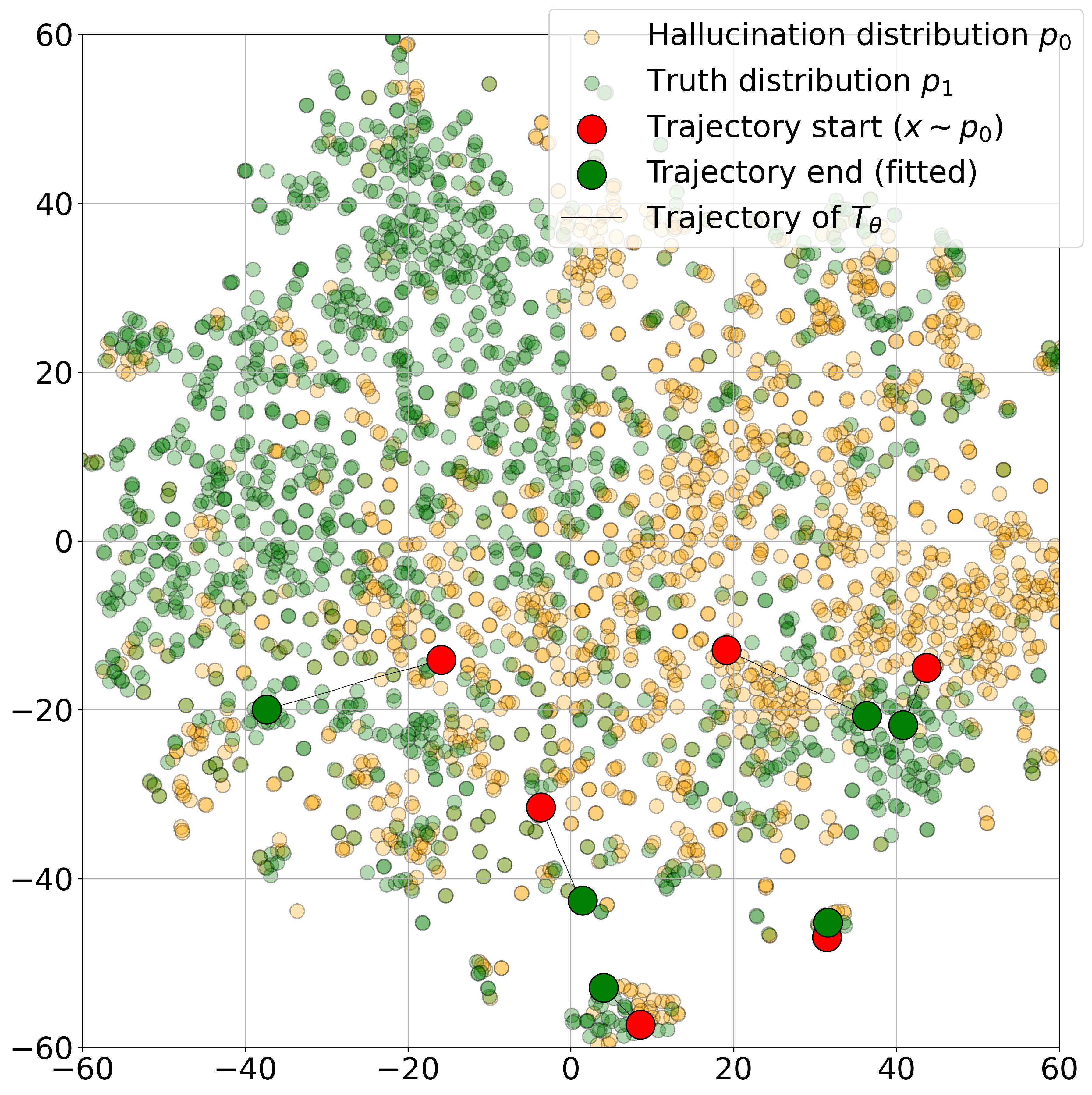} 
        \caption{Image-level, L2, Head 57, LLaVA-1.5-7B}
        \label{fig:image_level_ot_llava}
    \end{subfigure}
    \begin{subfigure}[b]{0.24\linewidth}
        \centering
        \includegraphics[width=\linewidth]{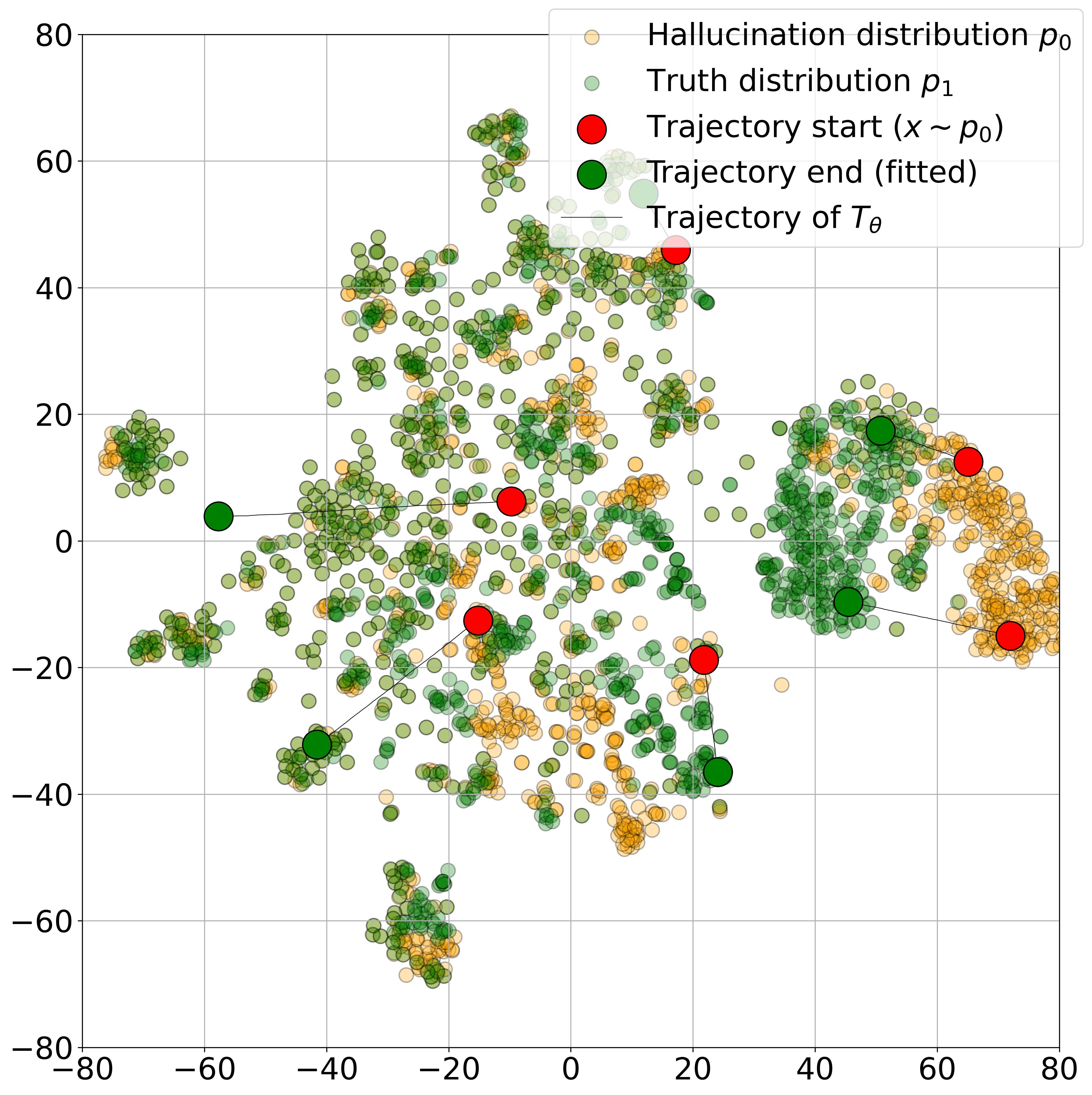} 
        \caption{Object-level, L14, Head 97, LLaVA-1.5-7B}
        \label{fig:object_level_ot_llava}
    \end{subfigure}
    \begin{subfigure}[b]{0.24\linewidth}
        \centering
        \includegraphics[width=\linewidth]{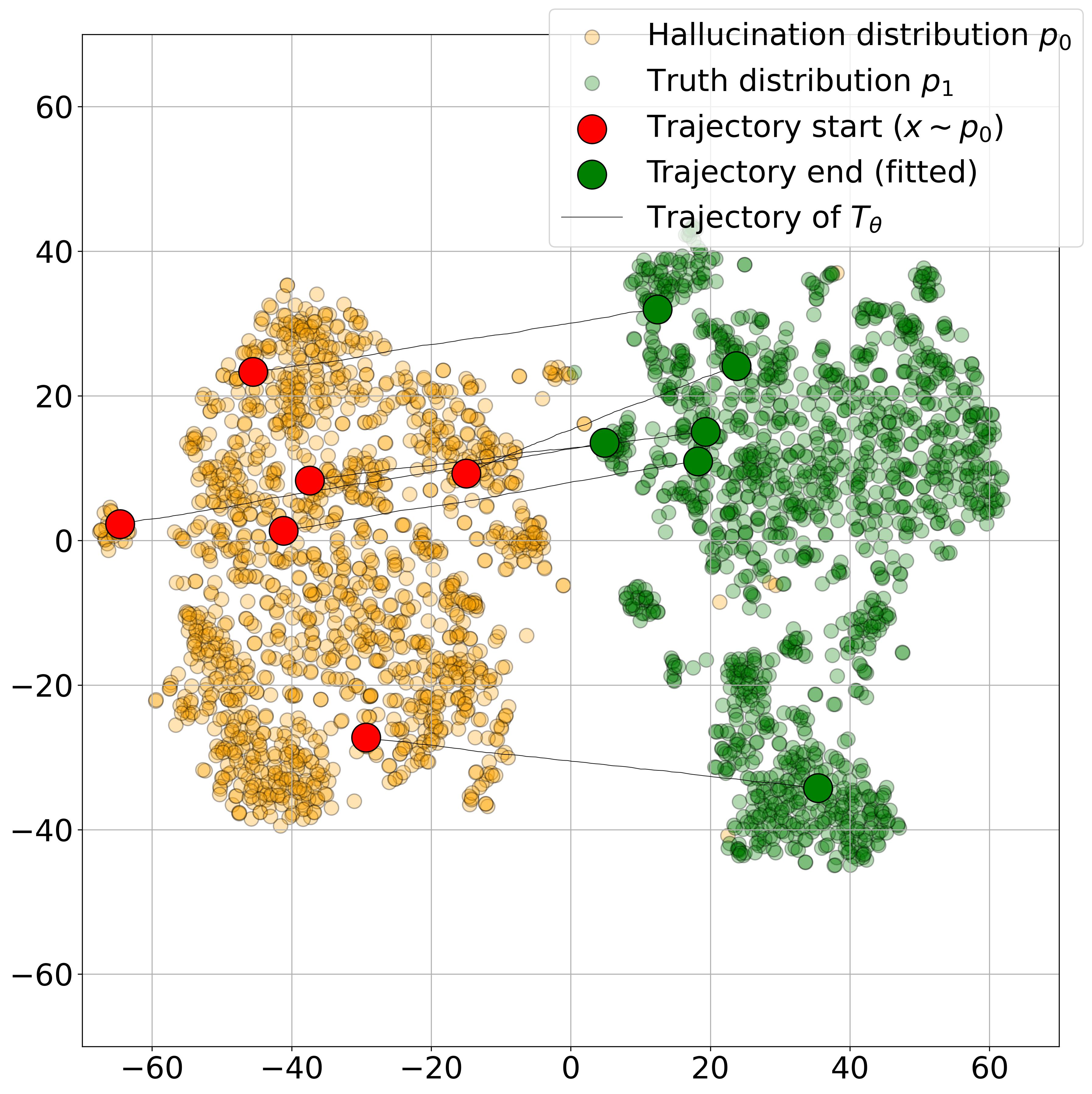} 
        \caption{Image-level, L17, Head 78, Qwen2.5-VL-7B}
        \label{fig:image_level_ot_qwen}
    \end{subfigure}
    \begin{subfigure}[b]{0.24\linewidth}
        \centering
        \includegraphics[width=\linewidth]{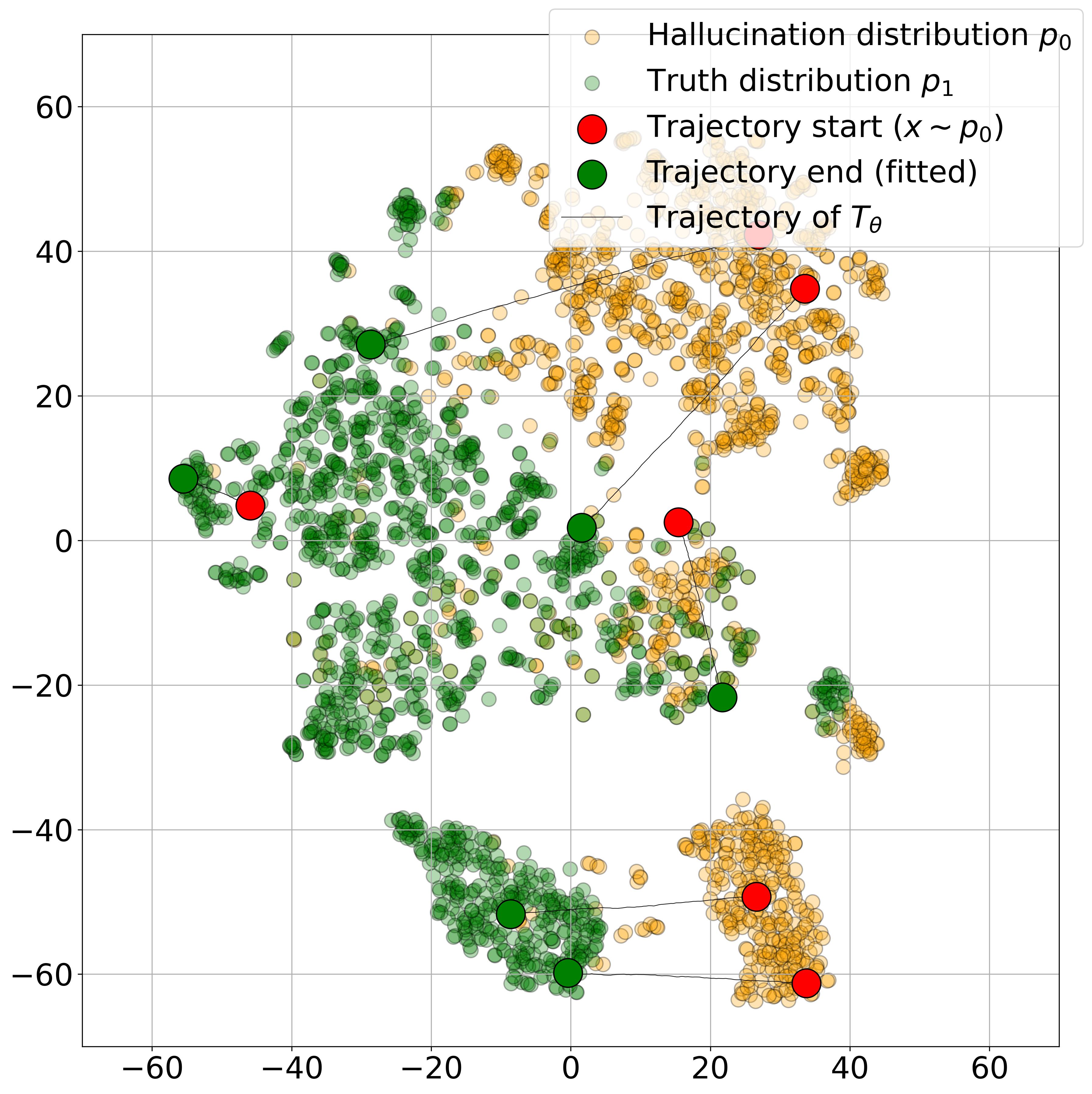} 
        \caption{Object-level, L14, Head 113, Qwen2.5-VL-7B}
        \label{fig:object_level_ot_qwen}
    \end{subfigure}
    \caption{A comparison of the hallucination and reliable manifolds in 
    LLaVA-1.5-7B and Qwen2.5-VL-7B is presented through t-SNE visualizations 
    at image and object levels, with transformations 
    between manifolds illustrated via SBP trajectories.}
    \label{fig:ot_plans}
\end{figure*}

While $\pi^*$ provides the optimal static mapping, it lacks temporal dynamics. 
We therefore introduce a time dimension $t \in [0,1]$ with $t=0$ and $t=1$ 
corresponding to $\mathbb{P}_{\text{hallu}}$ and $\mathbb{P}_{\text{fact}}$, 
respectively. Using a Wiener process $\mathbb{W}^\epsilon$ with 
diffusion coefficient $\sqrt{\epsilon}$ as prior, the SBP/EOT yields a 
diffusion process governed by:
\begin{align}
 \label{eq:sbp}
d\rva_t = \mathbf{g}^*(\rva_t, t)dt + \sqrt{\epsilon}  d\mathbf{W}_t, 
\end{align}
where the optimal drift $\mathbf{g}^*$ steers activations away from 
hallucinatory modes, as shown in Fig.~\ref{fig:ot_plans}.

The EOT-SBP equivalence establishes marginal consistency ($\pi^{\mathbb{T}^*} = \pi^*$) and reveals intermediate paths as Brownian bridges. This duality enables two mitigation strategies: static correction via $\pi^*$ for instantaneous remapping, and dynamic intervention using $\mathbf{g}^*(\rva, t)$ for progressive adjustment during inference.

\subsection{Hallucination-to-Truth Manifold Mapping}

The SBP/EOT solution to Eq.~(\ref{eq:eot}) admits a closed 
form~\cite{leonard2013survey,korotin2024light}:
\begin{align}
\label{eq:close_solution_static}
\pi^*(\rva_0, \rva_1) = u^*(\rva_0)  \exp\left(\langle \rva_0, \rva_1 \rangle / \epsilon\right) v^*(\rva_1),
\end{align}  
where $\rva_0 \equiv \rva_{\text{hallu}}$, $\rva_1 \equiv \rva_{\text{fact}}$, 
and $u^*$, $v^*$ incorporate hallucination and reliable Schr\"{o}dinger potentials. This yields the conditional distribution $\pi^*(\rva_1|\rva_0) \propto \exp(\langle \rva_0, \rva_1 \rangle / \epsilon) v^*(\rva_1)$ for transforming hallucinated activations. The correction path is governed by:
\begin{align} 
g^*(\rva, t) = \epsilon \nabla_{\rva} \log \int \mathcal{N}(\rva'|\rva,(1-t)\epsilon \mathbf{I}) \phi^*(\rva') d\rva',
\end{align}  
defining a smooth transport field that avoids semantic discontinuities.

Inspired by Korotin et al.~\cite{korotin2024light}, 
we approximate 
trustworthy potential $v^*$ via Gaussian mixtures:
\begin{align} \label{eq:v_theta}
v_\theta(\rva_1) = \sum_{i=1}^G \alpha_i \mathcal{N}(\rva_1 \mid r_i, \epsilon S_i),
\end{align}
with learnable $\theta = \{\alpha_i, r_i, S_i\}$. The conditional distribution becomes:
{\small
\begin{align}
\label{eq:a1_given_a0}
\pi_\theta(\rva_1 \mid \rva_0) = \frac{1}{c_\theta(\rva_0)} \sum_{i=1}^G \alpha_{i}(\rva_0) \mathcal{N}(\rva_1 \mid r_i + S_i \rva_0, \epsilon S_i),
\end{align}
where $\alpha_{i}(\rva_0) = \alpha_i \exp( (\rva_0^\top S_i \rva_0 + 2r_i^\top \rva_0)/2\epsilon )$ and $c_\theta(\rva_0) = \sum_{i=1}^G \alpha_{i}(\rva_0)$.

We minimize the KL divergence between $\pi^*$ and $\pi_\theta$ via:
$$\mathcal{L}(\theta) = \mathbb{E}_{\rva_0 \sim p_0}[\log c_\theta(\rva_0)] - \mathbb{E}_{\rva_1 \sim p_1}[\log v_\theta(\rva_1)],$$
optimized through mini-batch SGD. This approach enables distribution-level 
hallucination correction while preserving semantic fidelity and requiring 
only unpaired hallucination/trustworthy samples.

\subsection{Hallucination Mitigation Pipeline for MLLMs}  
Building on prior theory, as shown in Fig.~\ref{fig:framework} 
Schr\"{o}Mind implements a two-stage pipeline. 
First, we extract attention activations from all Transformer 
heads (Eq.~\ref{eq:transformer}) using both correct and hallucinated 
VQA pairs. These activations train logistic regression classifiers per head 
to distinguish hallucinated/factual outputs and enable selection of the top $H$ 
heads (ranked by classification accuracy) for intervention.

Second, for each selected head, we solve Eq.~(\ref{eq:eot}) and \
Eq.~(\ref{eq:sbp}) to compute an optimal transport plan and 
Schr\"{o}dinger bridge between activation distributions. This dynamically 
steers unknown tokens' activations toward factual manifolds:  
\begin{equation}
\rvx^{(k)} = \rvx^{(k-1)} + \sum_{m=1}^{M} \left[
\mathcal{T}_m^{k-1}\big( \rvx^{(k-1)} \big) \mapsto {\rva_1\vert}_{m=1}^{M}  \right]\mathbf{\Theta}_m^{k-1}, \nonumber
\end{equation}  
where this steering operation maps potentially hallucinated activations
$\mathcal{T}_m^{k-1}\big( \rvx^{(k-1)} \big)$
to hallucination-resistant 
distributions ${\rva_1\vert}_{m=1}^{M} $ via $\pi_\theta(\rva_1 \mid \rva_0)$ (Eq.~\ref{eq:a1_given_a0}).

The Schr\"{o}dinger bridge process (Eq.~\ref{eq:sbp}) is discretized as:  
\begin{equation}
\rva_{t+\Delta t} = \rva_t + g_\theta(\rva_t, t)\Delta t + 
\sqrt{\epsilon \Delta t} \, \xi, \quad \xi \sim \mathcal{N}(0, \mathbf{I}),
\end{equation}  
with precomputed drift $g_\theta$ guiding activations toward factual outputs. Mitigation strength is controlled by $t \in [0,1]$, where $t=0$ ($\rva_0$) implies no intervention and $t=1$ ($\rva_1$) full intervention.

Interventions occur at two granularities: (1) \textit{image-level} (whole-image 
perturbations), and (2) \textit{object-level} (object-specific perturbations). 
When both exist for a head, results are averaged; otherwise, the single applicable 
intervention is applied. As shown in Fig.~\ref{fig:ot_plans}, the distributions of hallucinated versus factual activations differ significantly across the two visual input settings, 
necessitating distinct hallucination mitigation trajectories.

\section{Experiments}
\label{sec:exp}

\subsection{Benchmarks and Evaluation Protocol}

\textbf{Polling-based Object Probing Evaluation (POPE).}  
We evaluate object hallucinations using POPE~\cite{li2023evaluating}, which 
employs 27K balanced yes/no queries (half with present/absent objects) 
from MS COCO~\cite{lin2014microsoft}, A-OKVQA~\cite{schwenk2022okvqa}, 
and GQA~\cite{hudson2019gqa}. Unlike caption-based methods, POPE
directly quantifies hallucination tendencies through three sampling strategies (random, popular, adversarial), reporting Accuracy and F1-score.

\textbf{Multimodal Model Evaluation (MME).}  
MME~\cite{fu2023mme} provides holistic assessment across 14 dimensions: 
10 perception tasks (object presence/quantity, spatial 
relations, color attributes) and 4 cognition tasks (commonsense reasoning (CSR), 
numerical calculation, text translation, program synthesis), with primary reliance on accuracy metrics.

\begin{table*}[!h]\small
   \caption{
Performance comparison of Schr{\"o}Mind and baselines on the POPE benchmark 
using the \colorbox{blue!15}{LLaVA-1.5-7B} model on three datasets: MSCOCO, A-OKVQA, and GQA. \textbf{Bold} and \underline{underlined} values show the best and second-best, respectively.
   }
 \label{tab:SchroMind_llava}
  \centering
  \setlength{\tabcolsep}{4pt}
	\begin{tabular}{|l|ccc|ccc|ccc|}
      \hline
	\multirow{2}{*}{Acc.$\uparrow$/F1$\uparrow$} & \multicolumn{3}{c|}{MS COCO} 
	&  \multicolumn{3}{c|}{A-OKVQA} &  \multicolumn{3}{c|}{GQA} 
	 \\ 
	   \cline{2-10} 
			& Random	& Popular & Adversarial & Random & Popular & Adversarial
				  & Random  & Popular
				  & Adversarial  
				  \\ 
              \hline 
              Regular~\cite{liu2023visual}   & 83.29/81.33 & 81.88/80.06 & 78.96/77.57 
              &  83.45/82.56 & 79.90/79.59 & 74.04/75.15 
              & 83.73/82.95 &  78.17/78.37 & 75.08/76.06 \\ 
              VCD~\cite{leng2024mitigating} &  87.73/87.16 & 85.38/85.06 &  80.88/81.3 
              &  86.15/86.34 & 81.85/82.82 & 74.97/77.73 
              & 86.65/86.99 & 80.73/82.24 &  76.09/78.78 \\ 
               M3ID~\cite{favero2024multi} & 86.00/86.00  & 82.83/83.72 & 77.70/79.66
               & 83.57/85.09 & 76.80/80.06  & 68.10/74.58
               & 82.83/84.62  & 72.83/77.58 &  68.13/74.78 \\ 
                OPERA~\cite{huang2024opera}  &  \underline{89.20}/\underline{88.81} & 86.64/\underline{86.62} & 81.24/81.38
               & 88.02/84.59 &  83.22/\underline{84.67} & 73.82/77.91
               &  88.13/88.91 & 79.27/82.11 &  75.00/78.71 \\ 
               AVISC~\cite{woo2024dont} & 87.93/87.88  & 84.33/84.96  & 77.53/79.64
               &  84.60/85.88 & 78.83/81.63  & 68.97/75.11
               &  85.00/86.45 & 74.80/79.17 & 69.20/75.58  \\ 
                  ICT~\cite{chen2024ict} & 89.1/88.48 & \underline{86.76}/86.40 & \underline{83.83}/\underline{83.84}
                  & \underline{89.3}/\underline{89.40}  & \underline{83.4}/84.45 & \underline{75.56}/\underline{78.68}
                  & \textbf{89.3}/\textbf{89.49} & \underline{80.86}/\underline{82.64}  & \underline{77.4}/\underline{80.11}\\ 
             \rowcolor{gray!30}   
                  Schr{\"o}Mind & \textbf{90.86}/\textbf{89.11} & 
              \textbf{87.1}/\textbf{87.48} & \textbf{85.43}/\textbf{84.48}
              & \textbf{90.83}/\textbf{90.77} & \textbf{85.63}/\textbf{85.42} & 
              \textbf{79.1}/\textbf{80.15}
              &  \underline{89.23}/\underline{89.22}  & \textbf{84.83}/\textbf{84.93}  
              &  \textbf{81.76}/\textbf{82.31} \\ 
  \hline
  \end{tabular}
\end{table*}

\begin{table*}[!h]\small 
   \caption{ 
Performance comparison between our Schr{\"o}Mind (SM) and previous SOTA (ICT) on 
the POPE benchmark using the \colorbox{blue!15}{Qwen2.5-VL-7B} model across 
MS COCO, A-OKVQA, and GQA datasets. Results are shown for image-level (w/o obj), 
object-level (w/o img), and combined interventions. Gray-highlighted entries denote 
our method.
 }
 \label{tab:SchroMind_ict_qwen2_5_vl_on_pope}
  \centering
  \setlength{\tabcolsep}{4pt}
	\begin{tabular}{|l|ccc|ccc|ccc|}
      \hline
	\multirow{2}{*}{Acc.$\uparrow$/F1$\uparrow$} & \multicolumn{3}{c|}{MS COCO} 
	&  \multicolumn{3}{c|}{A-OKVQA} &  \multicolumn{3}{c|}{GQA} 
	 \\ 
	   \cline{2-10} 
			& Random	& Popular & Adversarial & Random & Popular & Adversarial
				  & Random  & Popular
				  & Adversarial  
				  \\ 
          \hline 
                  Regular~\cite{bai2025qwen2} & 85.4/82.97 & 85.13/82.71 & 84.83/82.42
                  &  87.76/86.41 & 86.43/85.15  & 81.5/80.78
                  &  87.1/85.63 & 84/82.74 & 81.6/80.65 \\ 
              \hline 
                  ICT w/o obj& 86.4/84.31 & 86.3/84.27 & 85.7/83.53
                  &  89.93/89.15 & 87.46/86.76 & 82.5/82.54
                  &  88.06/87.01 & 83.9/83.23 & 81.63/81.17\\ 
     \rowcolor{gray!30}         
     SM w/o obj & 87.53/85.13 &	87.33/84.97	& 87.0/84.62
              & 90.3/89.08	& 88.56/87.43 &	83.56/83.04
              & 90.06/88.77 &	85.3/84.23 &	82.76/81.97 \\ 
              \hline
   ICT w/o img & 85.4/82.97 & 85.4/83.07 & 85.2/82.90
                  & 88.76/87.64 & 86.63/85.50 & 81.7/81.16
                  & 87.16/85.66 & 83.93/82.67  & 81.33/80.40 \\ 
   \rowcolor{gray!30}           SM w/o img 
   & 87.6/85.22 &	87.26/84.86	& 87.5/84.72
              & 90.03/88.77	& 88.4/87.19 &	83.63/83.05
              & 90.33/89.10	& 85.36/84.33	& 82.83/82.12 \\ 
  \hline
   ICT~\cite{chen2024ict} &  87.53/85.84  & 86.76/84.94 & 86.16/84.32
                  & 88.96/87.90 & 87.43/86.39  & 83.6/83.02
                  &  88.96/87.89 &  86.43/85.47 & 84.1/83.53\\ 
      \rowcolor{gray!30}       
      SM (ours) 
      &  87.93/85.68 &	87.6/85.33	& 87.33/86.07
              &  90.66/89.53	& 88.83/87.80	& 83.96/83.52
              & 90.3/89.09	& 87.53/86.52 &	84.93/84.18  \\ 
  \hline
  \end{tabular}
\end{table*}

\textbf{Experimental Setup.}  
We evaluate Schr\"{o}Mind on LLaVA-1.5-7B~\cite{liu2023visual} and 
Qwen2.5-VL-7B~\cite{bai2025qwen2}, comparing against VCD~\cite{liu2023visual}, OPERA, 
and ICT~\cite{chen2024ict}\footnote[1]{ICT results reproduced from official implementation: \url{https://github.com/THU-BPM/ICT}.}. 
Compared to ICT
which requires tuning two hyperparameters, 
our Schr\"{o}Mind relies solely on the top-$k$ attention heads hyperparameter.
Following ICT~\cite{chen2024ict}, we train logistic regression 
classifiers on 1,500 activation samples (image/object-level) to 
identify top-$k$ hallucination-sensitive heads.

\begin{figure}[th]
  \centering
  \hspace{-5mm}
  \includegraphics[width=1.0\linewidth]{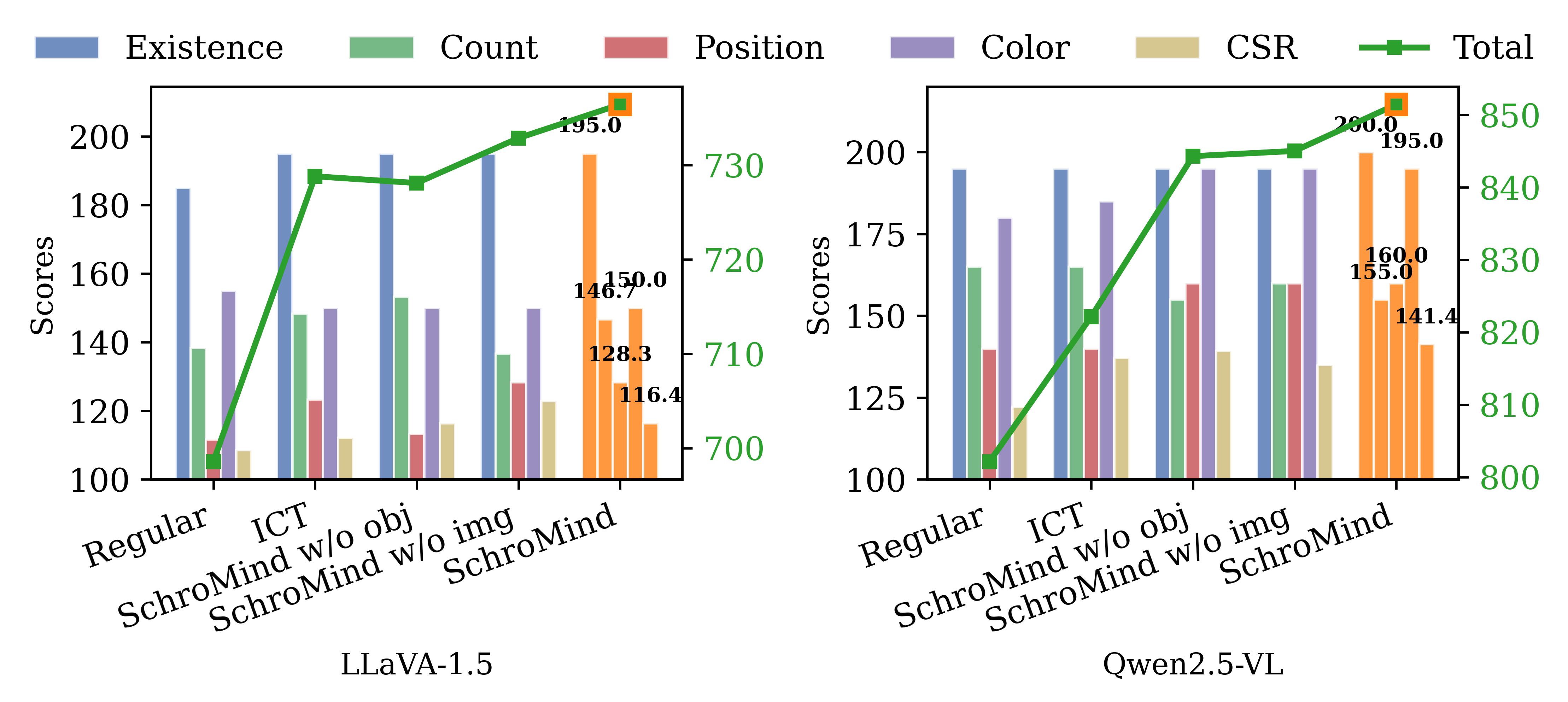}
  \hspace{-5mm}
  \caption{
 Schr{\"o}Mind outperforms prior SOTA models (ICT, regular LLaVA-1.5 and Qwen2.5-VL) 
 on MME, with 
gains in key areas: existence, position, counting, color perception, commonsense 
reasoning, and overall performance.  
  }
  \label{fig:mme_results}
  \end{figure}

\subsection{Results and discussion}
\label{sec:results}

As shown in Table \ref{tab:SchroMind_llava}, Schr\"{o}Mind achieves 
significant performance improvements over both the unmodified regular 
LLaVA-1.5-7B
model and the previous SOTA method ICT across all POPE datasets 
(MS COCO, A-OKVQA, and GQA). On challenging Adversarial questions, 
it attains accuracy gains of up to +6.68\% (GQA), +5.06\% (A-OKVQA), and 
+6.47\% (MS COCO) compared to the regular model. Against ICT, 
Schr\"{o}Mind leads in 17 out of 18 metrics, with notable improvements 
in adversarial settings: +3.54\% on A-OKVQA and +4.36\% on GQA. It 
is the only method consistently exceeding 80\% F1 on Adversarial 
questions across all datasets, highlighting its enhanced reliability in 
high-risk scenarios.

Results in Table~\ref{tab:SchroMind_ict_qwen2_5_vl_on_pope} 
further confirm these advantages across model architectures. 
With 
Qwen2.5-VL-7B, it maintains consistent leads,
 such as +1.17\% on MS COCO and +0.83\% on GQA Adversarial, and 
 achieves a new benchmark of 87.53\% on GQA Popular. These 
 results establish Schr\"{o}Mind as a universally superior
  framework for reducing hallucinations.

On the MME benchmark (Fig.~\ref{fig:mme_results}), Schr\"{o}Mind also 
outperforms both baseline and ICT methods. With LLaVA-1.5-7B, it reaches a 
total score of 736.42, exceeding ICT by +7.62 and Regular by +37.85, 
showing special strength in tasks of \textit{Position} and \textit{Count}. 
Using Qwen2.5-VL-7B, it sets 
a new SOTA total score of 851.42, leading in tasks of \textit{Color}
 (195.0), \textit{Position} (160.0), 
and \textit{CSR} (141.4).

\subsection{Ablation study}
\label{sec:ablation}

To evaluate component contributions in our Schr{\"o}Mind  framework, we 
conducted an ablation study on the POPE benchmark using Qwen2.5-VL-7B. We 
compared image-level intervention, object-level intervention, 
and their full integration.
As shown in Table~\ref{tab:SchroMind_ict_qwen2_5_vl_on_pope}, both 
interventions significantly outperform the baseline, with their combination 
yielding optimal performance. On MS COCO, individual interventions achieve 
competitive accuracy (87.0–87.6\% across subsets), while full Schr{\"o}Mind
delivers the strongest results—particularly in all settings. 
This pattern holds on A-OKVQA (e.g., 90.66\% Acc./89.53\% F1 on Random) and GQA, 
where full Schr{\"o}Mind achieves peak performance (90.3\% Acc. on 
Random; 87.53\% Acc./86.52\% 
F1 on Popular).
Crucially, the synergy between interventions consistently surpasses individual 
contributions across all datasets and question types, confirming their 
complementary role in hallucination suppression.

On the MME benchmark (Fig.~\ref{fig:mme_results}), the full model achieves 
the highest overall scores (LLaVA-1.5-7B: 736.42; Qwen2.5-VL-7B: 851.42), surpassing 
ablations 
by \textbf{+3.57–8.33} (LLaVA-1.5-7B) and \textbf{+6.42–7.14} (Qwen2.5-VL-7B). 
The variants 
show complementary strengths: \textit{w/o image} excels in \textit{Position} 
and \textit{CSR} but trails in \textit{Count}, while \textit{w/o object} shows 
the opposite. Only the complete model balances all capabilities, achieving 
top performance in key subsets such as \textit{Existence} (200.0), \textit{Color} 
(195.0), and \textit{CSR} (141.42), demonstrating that both intervention types 
are necessary for robust multimodal reasoning.

\section{Conclusion}
\label{sec:concl}

Without increasing computational cost, our proposed Schr{\"o}Mind 
method provides customized correction for hallucinatory attention activations 
generated by the Transformer in MLLMs by 
modeling the transition relationship between hallucinatory and credible 
attention through dynamic SBP (which is equivalent to static EOT), thereby 
constructing a correction scheme with the shortest overall migration distance 
and optimal cost-efficiency. This method effectively corrects attention without 
compromising the inherent capabilities of MLLMs acquired through extensive training 
on massive datasets. Experimental results across multiple datasets and benchmarks 
demonstrate that Schr{\"o}Mind achieves new state-of-the-art performance, significantly 
outperforming non-customized correction methods.




\bibliographystyle{IEEEbib}
\bibliography{strings,refs}

\end{document}